\documentclass[conference]{IEEEtran}
\usepackage{cite}
\usepackage{amsmath,amssymb,amsfonts}
\usepackage{algorithmic}
\usepackage{graphicx}
\usepackage{subcaption}
\usepackage{textcomp}
\usepackage{xcolor}
\usepackage{url}

\def\BibTeX{{\rm B\kern-.05em{\sc i\kern-.025em b}\kern-.08em
    T\kern-.1667em\lower.7ex\hbox{E}\kern-.125emX}}

\usepackage[draft=false]{defs-common}
\usepackage{defs-custom}

\begin{document}

\title{Active Learning for Classifying 2D Grid-Based Level Completability 
}

\author{\IEEEauthorblockN{Mahsa Bazzaz}
\IEEEauthorblockA{\textit{Khoury College of Computer Sciences} \\
\textit{Northeastern University}\\
Boston, USA \\
bazzaz.ma@northeastern.edu}
\and
\IEEEauthorblockN{Seth Cooper}
\IEEEauthorblockA{\textit{Khoury College of Computer Sciences} \\
\textit{Northeastern University}\\
Boston, USA \\
se.cooper@northeastern.edu}
}

\IEEEoverridecommandlockouts
\IEEEpubid{\makebox[\columnwidth]{979-8-3503-2277-4/23/\$31.00~\copyright2023 IEEE \hfill}
\hspace{\columnsep}\makebox[\columnwidth]{ }}
\maketitle
\IEEEpubidadjcol

\begin{abstract}
Determining the completability of levels generated by procedural generators such as machine learning models can be challenging, as it can involve the use of solver agents that often require a significant amount of time to analyze and solve levels. Active learning is not yet widely adopted in game evaluations, although it has been used successfully in natural language processing, image and speech recognition, and computer vision, where the availability of labeled data is limited or expensive. In this paper, we propose the use of active learning for learning level completability classification. Through an active learning approach, we train deep-learning models to classify the completability of generated levels for Super Mario Bros., Kid Icarus, and a Zelda-like game. We compare active learning for querying levels to label with completability against random queries. Our results show using an active learning approach to label levels results in better classifier performance with the same amount of labeled data.
\end{abstract}



\newcommand{\XFIGUREplaceholder}{
\begin{figure}[t]
\centering
\includegraphics[width=0.975\columnwidth]{figure/_placeholder}
\caption{\label{XFIGUREplaceholder} Placeholder figure.}
\end{figure}
}



\newcommand{\XTABLEplaceholder}{
\begin{table}[t]
\centering
\begin{tabular}{cc}
--- & --- \\
--- & ---
\end{tabular}
\caption{\label{XTABLEplaceholder} Placeholder table.}
\end{table}
}


\section{Introduction}
Completability is a key aspect of procedural level generation \cite{shaker_procedural_2016}. There are many approaches that try to ensure that generated levels are actually completable by the player. Some of these approaches try to integrate completability evaluation with the generation, and some follow the generate-and-test approach. 
However, these approaches often require the use of gameplay agents to attempt to solve a level. By contrast, machine learning approaches may be able to learn models to classify levels as completable or not more efficiently. While training such models can require large amounts of training examples, \emph{active learning} aims to reduce the amount of labeled data needed by selectively acquiring and labeling the most informative data points. It is particularly useful in scenarios where labeled data is scarce or expensive to obtain \cite{Reyes_strategy_2018}.
Therefore, we evaluated the use of active learning in training classifiers of completability of procedurally generated game levels of three games (Super Mario Bros., Kid Icarus, and a Zelda-like game).
Our work shows that using active learning to query labeled data from an oracle results in better efficiency of the learner, hence resulting in more efficient models that are less expensive as they need less labeled data.

\section{Related Work}
\subsection{Level Completability}
Procedural content generation via machine learning \cite{summerville_procedural_2018} enables the generation of video game levels that are aesthetically similar to human-authored examples. However, the generated levels can be unplayable without additional editing.  Multiple approaches like utilizing pathfinding agents and repairing generated levels have been used to ensure generated levels are completable. Cooper and Sarkar \cite{Cooper2020PathfindingAF} leveraged pathfinding agents to test levels for completability and perform level repair to fix generated levels that are unplayable. Also, Jain et al. \cite{jain2016autoencoders} used autoencoders to repair level content; related to the approach we present, they also used autoencoders to recognize the ``style'' of levels. Furthermore, Zhang et al. \cite{Zhang2020VideoGL} constructed a framework to generate levels using an adversarial network and repair them using a mixed-integer linear program with completability constraints.

Other approaches aim to only generate completable levels.  Font et al. \cite{font2016constrained} used fitness evaluation to guarantee the solvability of the levels by construction rather than by generate-and-test. Furthermore, Nelson and Smith \cite{nelson_asp_2016} use a constraint-based generator to create a reference solution along with the level design for ensuring the generation of playable maps. Cooper \cite{cooper_sturgen_2023} presented a constraint-based approach to level generation for 2D tile-based games that simultaneously generates an example playthrough of the level, demonstrating its completability.
\subsection{Active Learning}
Active learning is a machine learning technique that aims to reduce the amount of labeled data required for training by selectively choosing the most informative examples for labeling \cite{settles_active_2012}. The idea is to train a model on a small subset of labeled data, and then iteratively select a small set of unlabeled samples, ask an oracle (e.g. a human expert or another algorithm) to label those examples, and incorporate the newly labeled data into the training set. The algorithm then re-trains the model on the expanded labeled set and repeats the process until the desired performance is achieved. When the learner has all the unlabeled samples available before the start of the learning process, then the learning is known as pool-based \cite{kumar_active_2020}. Active learning can be particularly useful in scenarios where labeled data is expensive or difficult to obtain. By selectively labeling the most informative examples, active learning can improve the accuracy of the model while minimizing the cost of labeling. 

In classification tasks, active learning can be used with the goal of minimizing the labeling effort while maximizing the classification performance \cite{monarch2021human}.  Active learning can be applied to various classification problems, such as image classification, text classification, and sentiment analysis. For instance, Cao et al. \cite{Cao_image_classification_2020} used an active deep learning approach for image classification that achieves better performance with fewer labeled samples. Moreover, Yao et al. \cite{Yao_facial_2021} used 
active learning to reduce the cost of labeling samples and improve the training speed to classify human facial expressions. Additionally, Flores et al. \cite{Flores_Text_2021} used active learning in biomedical text classification algorithms, and were able to lessen the cost of manual annotation by specialized professionals.
\subsection{Active Learning in Games}
In games, active learning has been used in playtesting. For instance, Zook et al. \cite{zook_automatic_2014}, used active learning to formalize and automate a subset of playtesting goals and reduce the amount of playtesting needed to choose the optimal set of game parameters. Also, Normoyle et al. \cite{Normoyle_metric_2021} used active learning to collect player metric data more efficiently for improving the gameplay. Moreover, \cite{shi_learning_2018} used active learning for classifying game segments quality.
\section{Method}

\subsection{Goal}
The goal of our approach is to learn a classifier of the completability of levels for a game, and use active learning to improve the performance of the classifier. In this study, all the datasets were labeled in advance, and asking the oracle for the label is just fetching the correct label. We used pre-labeled datasets to focus on the performance of active learning without the implications of using an external oracle.
\subsection{Datasets}
\begin{figure}
    \centering
    \includegraphics[width=0.45\textwidth]{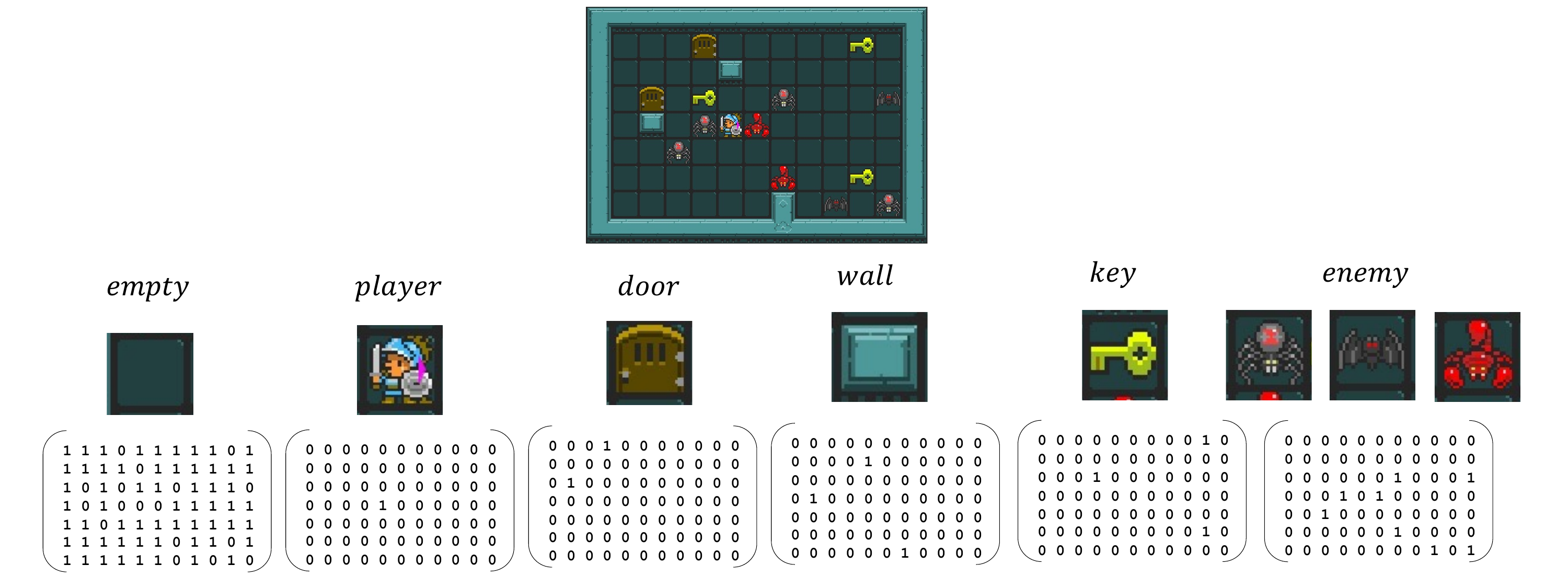}
    \caption{Data representation for a Zelda level.}
    \label{fig:legend_of_zelda_rep}
\end{figure}
We used three datasets of different game levels for our experiments. 
    
    We used a VGDL version of Zelda generated by \cite{ZAFAR_level_2020}. Levels provided in this dataset were labeled as playable and unplayable, therefore we used these labels as the oracle to ask for labels. After processing the level images into tiles, the tiles along the edge were discarded, resulting in $7\times11$ tile levels.
    
    We used level segments ($14\times 25$) of the Super Mario Bros. video game and level segments ($25\times 16$) of the Kid Icarus video game, generated by \cite{biemer2021gram}, and based on levels in the Video Game Level Corpus \cite{VGLC}.
    These levels were not labeled, but using the fitness values calculated for each level segment in previous work, we separated levels into completable and uncompletable.

For all datasets, we constructed a binary matrix for each unique tile type, including one-hot encoding of the tile type for each tile of the map, meaning the value 1 indicates the presence of a tile of that category and the value 0 indicates the absence of a tile of that category. Figure \ref{fig:legend_of_zelda_rep} shows binary arrays generated for representing a level.
The initial datasets were imbalanced regarding completability, so we used random under-sampling to balance them. 
This resulted in a dataset of 3100 levels for Super Mario Bros., 4490 levels for Kid Icarus, and 1508 levels for Zelda.

\subsection{Active Learning}
\begin{figure}
    \centering
    \includegraphics[width=0.45\textwidth]{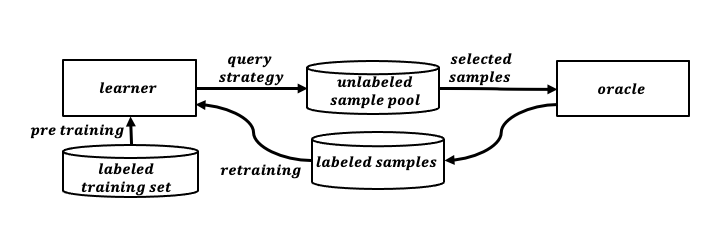}
    \vspace{-15pt}
    \caption{The pool-based active learning cycle.}
    \label{fig:active_learning}
\end{figure}
Active Learning aims to
achieve good accuracy with fewer training samples. There are several ways that a learner could be able to ask queries, one of which is pool-based sampling.
Figure \ref{fig:active_learning} represents the pool-based active learning that assumes that there is a small starting set of ``pre-training'' labeled data and a large pool of unlabeled data available.

For choosing queries from the unlabeled pool, one approach is \emph{uncertainty sampling}. Uncertainty sampling needs the classifier to output the probabilities of each class.  There are several uncertainty sampling techniques (least confident sampling, margin sampling, and entropy sampling), but for binary classification, all these strategies are monotonic functions of one another leading to similar performances \cite{settles_active_2012}. In this study, we used margin sampling as a representative of uncertainty sampling strategies. When querying for unlabeled samples with margin sampling, this strategy selects the sample with the smallest margin (i.e. difference in certainty between the two classes), since the smaller the decision margin is, the more unsure the decision.

On the other hand, machine learning models generally use random sampling for training. Thus, we compared the impact of using active learning uncertainty sampling methods to two baselines. First, \emph{random sampling}, which uses the same pool-based iterative sampling approach, but simply chooses each sample randomly.  Second, \emph{passive learning}, which uses all the training data available (in each fold) to train the model.

\begin{figure*}
     \centering
     \begin{subfigure}[h]{0.32\textwidth}
         \centering
         {\includegraphics[width=\textwidth]{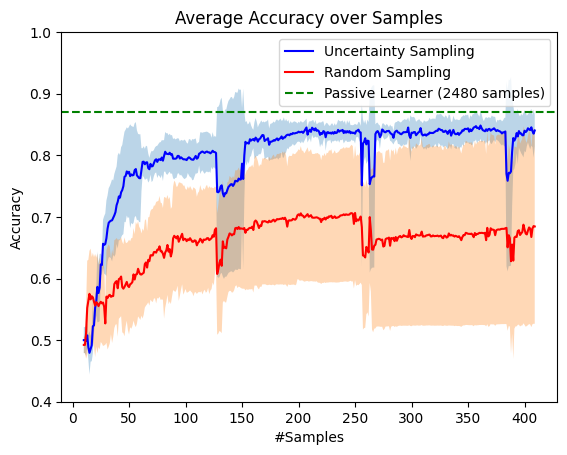}}
         \caption{Super Mario Bros.}
         \label{fig:accuracy-mario}
     \end{subfigure}%
     \hfill
     \begin{subfigure}[h]{0.32\textwidth}
         \centering
         {\includegraphics[width=\textwidth]{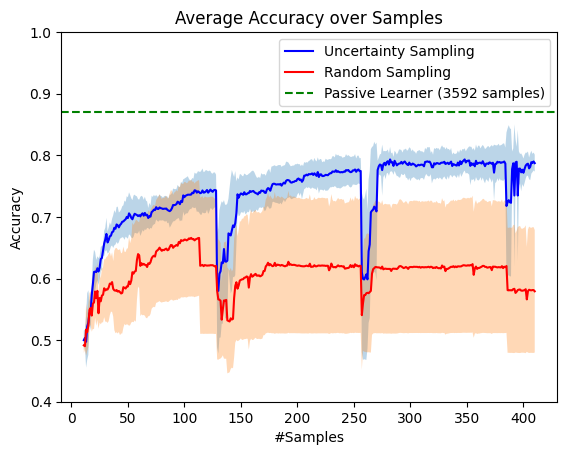}}
         \caption{Kid Icarus}
         \label{fig:accuracy-icarus}
     \end{subfigure}%
     \hfill
     \begin{subfigure}[h]{0.32\textwidth}
         \centering
         {\includegraphics[width=\textwidth]{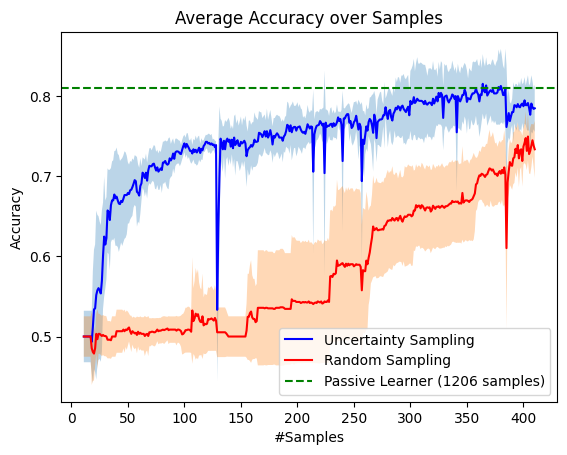}}
         \caption{Zelda}
         \label{fig:accuracy-zelda}
     \end{subfigure}
     \caption{Comparison of mean accuracy over number of samples. Shading shows standard deviation. The dotted line shows the passive learner, which is trained on all samples in each fold. In addition to the queries, sampling learners had an additional 10 samples from the pre-training.}
     \label{fig:accuracy}
\end{figure*}

\begin{table}
\newcommand{\z}{\hphantom{0}}
  \centering
  \caption{Average false discovery rate (FDR) and false omission rate (FOR) from final models (as percentage).}
  \label{tab:confusion-matrix}
  \begin{tabular}{|l|c|c|c|}
    \hline
     & Super Mario Bros. & Kid Icarus & Zelda \\
    \hline
    FDR Uncertainty & 23.7 & 32.5 & 20.6 \\
    \hline
    FOR Uncertainty & \z8.1 & \z9.9 & 14.8 \\
    \hline
    \hline
    FDR Random & 61.4 & 99.2 & 48.6 \\
    \hline
    FOR Random & \z1.7 & \z0.7 & \z4.6 \\
    \hline
    \hline
    FDR Passive & \z9.3 & \z8.8 & 13.6 \\
    \hline
    FOR Passive & 14.5 & 12.8 & 19.5 \\
    \hline
  \end{tabular}
\end{table}

\subsection{Training}
We did all implementation of the models and training in Google Colab~\cite{bisong2019google} and all the code is available in GitHub\footnote{https://github.com/MahsaBazzaz/level-completabilty-x-active-learning}. We used binary convolutional neural networks as base classifiers. These models use three convolutional layers to extract features from the input data, each followed by max pooling layers to reduce the dimensionality of the feature maps, and ReLU activation function to introduce non-linearity and help the model learn more complex patterns in the data. At last, fully connected layers are used to produce the final classification output. This is a 2D output passed through a softmax activation function resulting in a probability value for each class.
The number of input channels of each model is different, as it is the number of one-hot matrices needed to represent the different tiles present in levels: 8 for Super Mario Bros. and 6 for Zelda and Kid Icarus. Binary cross entropy loss function is used to quantify the dissimilarity between predicted and true class distributions. Furthermore, Adam optimizer provides parameters of the model that need to be optimized with a learning rate of 0.001 and weight decay of 0.001. Each training used 50 epochs.

We split $20\%$ of the dataset into a test dataset and the rest into a pool dataset. The test set is a completely independent set of data that is used to evaluate the performance of the model. We used 5-fold cross-validation to have a more reliable estimate of the model's performance than a single split of the data into pool and validation sets.

Training the sampling classifiers consists of two phases: In the pre-training phase, a batch of labeled training data is introduced to the model in order to bootstrap the learner. After that, the model starts to query for samples and then re-trains on the newly fetched labeled data. In order to be able to compare uncertainty sampling and random sampling, we used the same number of queries (each query fetching a single sample).

In order to compare uncertainty sampling and random sampling, we started the pre-training by randomly selecting 10 samples from the pool. After that, we queried 400 samples one at a time to monitor the performance of each sampling strategy with the least minimum number of samples; this number was selected as it was found to achieve close to passive learner performance in preliminary evaluations.
When querying with uncertainty sampling, the selected samples can be very similar to each other, which results in little information being added to update the classifier. Therefore, In each step before querying, in order to introduce randomness to sampling, we selected 50\% of the pool and selected a sample from this subset. This not only provides sampling strategy diversity, but also reduces the chances of outlier data getting selected by the sampler. To account for randomness in the sampling learners, we ran 5 trials of each fold, and used average performance across these trials for all folds (thus 25 trials per sampling strategy) for comparison; the passive learner used one trial per fold (5 trials overall).


\subsection{Metrics}
We used accuracy over the number of queries of different learners to compare them. As we balanced the datasets, accuracy is a useful metric to get insights into the performance of sampling strategies\cite{Reyes_strategy_2018}. For each trial, we calculated the accuracy of each model after the learner was done learning the new point of labeled data it has received from the oracle and appended that to the accuracy history. By doing so, after calculating the mean of accuracy after each query in all trials, we observe which sampling strategy has better performance after receiving data labeled from the oracle. Furthermore, because the goal of the classifier is to provide levels that are completable, it is important to consider that of all the levels that the classifier labeled as completable, how many are actually completable. So we monitored the average \emph{false discovery rate} (FDR), $\frac{false~positives}{total~positives}$, and \emph{false omission rate} (FOR), $\frac{false~negatives}{total~negatives}$, at the end of each trial. We also observed the average training time of each fold for sampling learners and compared it to the passive learner to get an understanding of the impact of active learning strategies on training time.

\section{Discussion}
As Figure~\ref{fig:accuracy} shows, our results demonstrate that with the same number of instances and model hyperparameters, the learner that uses an uncertainty sampling strategy achieves higher accuracy with less labeled data than random sampling. We see that the accuracy of the uncertainty sampling increases rapidly at the beginning, but there are diminishing returns as the number of samples already incorporated grows. In the given number of samples, uncertainty sampling can approach very close to the accuracy of the passive learner but does not necessarily achieve it.

If the classifier were part of a larger generate-and-test system, levels classified as playable could be passed on to an agent to confirm, and those classified as unplayable quickly be discarded without test. As table \ref{tab:confusion-matrix} shows, the sampling learners have a higher FDR (unplayable levels classified as playable, that would fail an agent test) than FOR (playable levels classified as unplayable that would be incorrectly discarded).

Regarding the training time, In super Mario Bros., each trial of passive learner took 13.6 seconds to train, active learning with uncertainty sampling took 8 minutes, and random sampling took 5.5 minutes. These numbers for Kid Icarus are 27 seconds, 15.8 minutes, and 11.5 minutes, and for Zelda 3 seconds, 7 minutes, and 2.5 minutes. In future work, we would like to include the timing of oracle labeling instances, and make a thorough comparison.

\section{Conclusion}

Our study shows that there is potential for active learning to reduce the number of samples needed in classifying levels, though more work is needed to look at the overall impact on the speed of the system. In this study, we only looked at classifier performance; in the future, we would like to integrate our model with completability agents as an oracle rather than pre-labeled data to evaluate the impact on timing (considering the time to train the classifier). Moreover, in this study, we balanced the datasets, but in the future, we would like to consider how uncertainty sampling can be used in imbalanced datasets.
Different sampling methods may apply if we expand beyond completability to multi-label classification, for example, discrete ranges of difficulty. Finally, in the future, we would like to explore the impact this potential of active learning may have on generate-and-test approaches by reducing the need to run agents.

\bibliography{refs-manual}

\begin{thebibliography}{10}
\providecommand{\url}[1]{#1}
\csname url@samestyle\endcsname
\providecommand{\newblock}{\relax}
\providecommand{\bibinfo}[2]{#2}
\providecommand{\BIBentrySTDinterwordspacing}{\spaceskip=0pt\relax}
\providecommand{\BIBentryALTinterwordstretchfactor}{4}
\providecommand{\BIBentryALTinterwordspacing}{\spaceskip=\fontdimen2\font plus
\BIBentryALTinterwordstretchfactor\fontdimen3\font minus
  \fontdimen4\font\relax}
\providecommand{\BIBforeignlanguage}[2]{{%
\expandafter\ifx\csname l@#1\endcsname\relax
\typeout{** WARNING: IEEEtran.bst: No hyphenation pattern has been}%
\typeout{** loaded for the language `#1'. Using the pattern for}%
\typeout{** the default language instead.}%
\else
\language=\csname l@#1\endcsname
\fi
#2}}
\providecommand{\BIBdecl}{\relax}
\BIBdecl

\bibitem{shaker_procedural_2016}
N.~Shaker, J.~Togelius, and M.~J. Nelson, \emph{Procedural content generation
  in games}.\hskip 1em plus 0.5em minus 0.4em\relax Springer, 2016.

\bibitem{Reyes_strategy_2018}
O.~Reyes, A.~H. Altalhi, and S.~Ventura, ``Statistical comparisons of active
  learning strategies over multiple datasets,'' \emph{Knowledge-Based Systems},
  vol. 145, pp. 274--288, 2018.

\bibitem{summerville_procedural_2018}
A.~Summerville, S.~Snodgrass, M.~Guzdial, C.~Holmgård, A.~K. Hoover,
  A.~Isaksen, A.~Nealen, and J.~Togelius, ``Procedural {Content} {Generation}
  via {Machine} {Learning} ({PCGML}),'' \emph{IEEE Transactions on Games},
  vol.~10, no.~3, pp. 257--270, Sep. 2018.

\bibitem{Cooper2020PathfindingAF}
S.~Cooper and A.~Sarkar, ``Pathfinding agents for platformer level repair,'' in
  \emph{AIIDE Workshops}, 2020.

\bibitem{jain2016autoencoders}
R.~Jain, A.~Isaksen, C.~Holmg{\aa}rd, and J.~Togelius, ``Autoencoders for level
  generation, repair, and recognition,'' in \emph{Proceedings of the ICCC
  workshop on computational creativity and games}, vol.~9, 2016.

\bibitem{Zhang2020VideoGL}
H.~Zhang, M.~C. Fontaine, A.~K. Hoover, J.~Togelius, B.~N. Dilkina, and
  S.~Nikolaidis, ``Video game level repair via mixed integer linear
  programming,'' in \emph{Artificial Intelligence and Interactive Digital
  Entertainment Conference}, 2020.

\bibitem{font2016constrained}
J.~M. Font, R.~Izquierdo, D.~Manrique, and J.~Togelius, ``Constrained level
  generation through grammar-based evolutionary algorithms,'' in
  \emph{Applications of Evolutionary Computation: 19th European Conference,
  EvoApplications 2016, Porto, Portugal, March 30--April 1, 2016, Proceedings,
  Part I 19}.\hskip 1em plus 0.5em minus 0.4em\relax Springer, 2016, pp.
  558--573.

\bibitem{nelson_asp_2016}
M.~J. Nelson and A.~M. Smith, ``{ASP} with applications to mazes and levels,''
  in \emph{Procedural {Content} {Generation} in {Games}}, N.~Shaker,
  J.~Togelius, and M.~J. Nelson, Eds.\hskip 1em plus 0.5em minus 0.4em\relax
  Springer International Publishing, 2016, pp. 143--157.

\bibitem{cooper_sturgen_2023}
S.~Cooper, ``{Sturgeon-MKIII}: Simultaneous level and example playthrough
  generation via constraint satisfaction with tile rewrite rules,'' in
  \emph{Proceedings of the 18th International Conference on the Foundations of
  Digital Games}, 2023.

\bibitem{settles_active_2012}
B.~Settles, \emph{Active {{Learning}}}.\hskip 1em plus 0.5em minus 0.4em\relax
  {Springer International Publishing}, 2012.

\bibitem{kumar_active_2020}
P.~Kumar and A.~Gupta, ``Active learning query strategies for classification,
  regression, and clustering: a survey,'' \emph{Journal of Computer Science and
  Technology}, vol.~35, pp. 913--945, 2020.

\bibitem{monarch2021human}
R.~M. Monarch, \emph{Human-in-the-Loop Machine Learning: Active learning and
  annotation for human-centered AI}.\hskip 1em plus 0.5em minus 0.4em\relax
  Simon and Schuster, 2021.

\bibitem{Cao_image_classification_2020}
X.~Cao, J.~Yao, Z.~Xu, and D.~Meng, ``Hyperspectral image classification with
  convolutional neural network and active learning,'' \emph{IEEE Transactions
  on Geoscience and Remote Sensing}, vol.~58, no.~7, pp. 4604--4616, 2020.

\bibitem{Yao_facial_2021}
L.~Yao, Y.~Wan, H.~Ni, and B.~Xu, ``Action unit classification for facial
  expression recognition using active learning and svm,'' \emph{Multimedia
  Tools and Applications}, vol.~80, 07 2021.

\bibitem{Flores_Text_2021}
C.~A. Flores, R.~L. Figueroa, and J.~E. Pezoa, ``Active learning for biomedical
  text classification based on automatically generated regular expressions,''
  \emph{IEEE Access}, vol.~9, pp. 38\,767--38\,777, 2021.

\bibitem{zook_automatic_2014}
A.~Zook, E.~Fruchter, and M.~O. Riedl, ``Automatic playtesting for game
  parameter tuning via active learning,'' in \emph{Proceedings of the 9th
  {{International Conference}} on the {{Foundations}} of {{Digital Games}}},
  2014.

\bibitem{Normoyle_metric_2021}
A.~Normoyle, J.~Drake, M.~Likhachev, and A.~Safonova, ``Game-based data capture
  for player metrics,'' \emph{Proceedings of the AAAI Conference on Artificial
  Intelligence and Interactive Digital Entertainment}, vol.~8, no.~1, pp.
  44--50, Jun. 2021.

\bibitem{shi_learning_2018}
P.~Shi and K.~Chen, ``Learning constructive primitives for real-time dynamic
  difficulty adjustment in super mario bros,'' \emph{IEEE Transactions on
  Games}, vol.~10, no.~2, pp. 155--169, Jun. 2018.

\bibitem{ZAFAR_level_2020}
A.~Zafar, H.~Mujtaba, and M.~O. Beg, ``Search-based procedural content
  generation for {GVG-LG},'' \emph{Applied Soft Computing}, vol.~86, p. 105909,
  2020.

\bibitem{biemer2021gram}
C.~Biemer, A.~Hervella, and S.~Cooper, ``Gram-elites: N-gram based
  quality-diversity search,'' in \emph{Proceedings of the 16th International
  Conference on the Foundations of Digital Games}, 2021.

\bibitem{VGLC}
A.~J. Summerville, S.~Snodgrass, M.~Mateas, and S.~Ontañón, ``The {VGLC}: The
  video game level corpus,'' in \emph{Proceedings of the 7th Workshop on
  Procedural Content Generation}, 2016.

\bibitem{bisong2019google}
E.~Bisong, ``Google colaboratory,'' \emph{Building machine learning and deep
  learning models on {Google} {Cloud} {Platform}: a comprehensive guide for
  beginners}, pp. 59--64, 2019.

\end{thebibliography}
\bibliographystyle{IEEEtran}

\end{document}